\documentclass{article}

\usepackage[nonatbib,preprint]{neurips_2025}

\usepackage{pifont}  
\usepackage[utf8]{inputenc} 
\usepackage[T1]{fontenc}    
\usepackage{hyperref}       
\usepackage{url}            
\usepackage{booktabs}       
\usepackage{amsfonts}       
\usepackage{nicefrac}       
\usepackage{microtype}      
\usepackage{xcolor}         
\usepackage[most]{tcolorbox}
\usepackage{subcaption}
\usepackage{float}
\usepackage{textcomp}
\usepackage{gensymb}
\usepackage{enumitem}
\usepackage{amsmath}
\usepackage{graphicx}
\usepackage{multirow}

\usepackage{listings}
\usepackage{pgfplots}
\pgfplotsset{compat=1.18}
\usepackage{adjustbox}

\hypersetup{colorlinks=true,linkcolor=blue,citecolor=blue,urlcolor=blue}

\newcommand{\std}[2]{#1 {\tiny$\pm$ #2}}

\title{Stable Prediction of Adverse Events \\ in Medical Time-Series Data}

\author{
    Mayank Keoliya\thanks{Equal Contribution}  \thanks{Corresponding authors: \texttt{seewon@seas.upenn.edu, mkeoliya@seas.upenn.edu}} , Seewon Choi\footnotemark[1]  \footnotemark[2], Rajeev Alur, Mayur Naik, Eric Wong\\
    University of Pennsylvania \\
    {\texttt{\{mkeoliya,seewon,alur,mnaik,exwong\}@seas.upenn.edu}}
}

\begin{document}

\maketitle
\setcounter{footnote}{0}

\begin{abstract}
Early event prediction (EEP) systems continuously estimate a patient's \textit{imminent risk} to support clinical decision-making. 
For bedside trust, risk trajectories must be accurate and temporally stable, shifting only with new, relevant evidence. 
However, current benchmarks (a) ignore stability of risk scores and (b) evaluate mainly on tabular inputs, leaving trajectory behavior untested. 
To address this gap, we introduce \textbf{CAREBench}, an EEP benchmark that evaluates deployability using multi-modal inputs—tabular EHR, ECG waveforms, and clinical text—and assesses temporal stability alongside predictive accuracy.
We propose a stability metric that quantifies short-term variability in per-patient risk and penalizes abrupt oscillations based on local-Lipschitz constants.
CAREBench spans six prediction tasks such as sepsis onset and compares classical learners, deep sequence models, and zero-shot LLMs. 
Across tasks, existing methods, especially LLMs, struggle to jointly optimize accuracy and stability, with notably poor recall at high-precision operating points. 
These results highlight the need for models that produce evidence-aligned, stable trajectories to earn clinician trust in continuous monitoring settings. (Code: \url{https://github.com/SeewonChoi/CAREBench}.)
\end{abstract}

\section{Introduction}

Predicting the occurrence of future \emph{events}---rather than forecasting the future value of a signal---from past signals is known as early event prediction (EEP). 
EEP underpins high-stakes decisions across healthcare \cite{muralitharan2021ews}, finance \cite{shi2019bankruptcy}, and manufacturing \cite{fernandes2022industrial}. 
We focus on the clinical setting, where models estimate the near-term risk of adverse events such as sepsis \cite{shashikumar2021sepsis, shashikumar2025sepsis, yeche2023smoothing} and cardiac arrest \cite{hyland2020cardiac, kwon2018cardiac}.

Clinical EEP is uniquely challenging. 
Patient records are \emph{multi-modal}, \emph{irregularly sampled}, and \emph{sparsely observed}. 
Structured tabular EHR features, unstructured clinical text, and continuous waveforms arrive at different cadences with varying noise and clinical salience \cite{wornow2025contextcluesevaluatinglong, yeche2023smoothing, shukla2021irregular}. 
For example, in MC-MED dataset \cite{mc-med}, ECG is recorded continuously, vitals are charted every minute, and lab tests such as sodium are typically measured every two hours.
Furthermore, there are over 300 different types of lab tests.
Prior works showed that tree models perform well on tabular data \cite{burger2024foundationmodelscriticalcare, mc-bec}, sequence models such as Mamba are suitable for capturing relationships across sparse and long-horizon data \cite{wornow2025contextcluesevaluatinglong}, and LLMs leverage clinical text well \cite{lovon20204mimicllm, ma2025memorizerankelevatinglarge}. 
However, it is unclear which architecture best handles the combination of the three modalities.
In addition, events such as sepsis are rare, making it challenging to balance precision and recall. 

\textbf{A need for a multi-modal benchmark.} Existing clinical benchmarks emphasize tabular EHR inputs and report \emph{per-window} metrics, making it difficult to compare methods developed for continuous monitoring on multi-modal data.
As summarized in Table~\ref{tab:benchmarks}, prior suites \cite{vandewater2024yaib, burger2024foundationmodelscriticalcare, yin2025mimic-ext} either lack multi-modality or omit evaluation beyond per-window discrimination, limiting insight into the behavior on full patient trajectories.

\newcommand{\cmark}{{\checkmark}} 
\newcommand{\xmark}{{\ding{55}}} 
\begin{table}[t]
\centering
\renewcommand{\arraystretch}{1.2}
\caption{Comparison of CAREBench with other clinical benchmarks.}
\label{tab:benchmarks}
\vspace{3mm}
\setlength{\tabcolsep}{4pt} 
\begin{adjustbox}{max width=\textwidth}
\begin{tabular}{l c c c c c}
\toprule
\textbf{Benchmark} & 
\textbf{\# Tasks} & 
\textbf{EEP} & 
\textbf{Multi-modal} & 
\textbf{Evaluation} & 
\textbf{Stability} \\
\midrule
YAIB \cite{vandewater2024yaib} & 2 & \cmark & \xmark~(EHR) & Per-window & \xmark \\
Burger et al \cite{burger2024foundationmodelscriticalcare} & 4 & \cmark & \xmark~(EHR) & Per-window & \xmark \\
MIMIC-III-Ext \cite{yin2025mimic-ext} & 4 & \xmark & \xmark~(EHR) & Per-window & \xmark \\
\textbf{CAREBench} & 6 & \cmark & \textbf{\cmark~(EHR, Waveform, Text)} & 
\textbf{Per-patient} & \cmark \\
\bottomrule
\end{tabular}
\end{adjustbox}
\end{table}

\begin{figure}[t]
    \centering
    \includegraphics[width=0.94\linewidth]{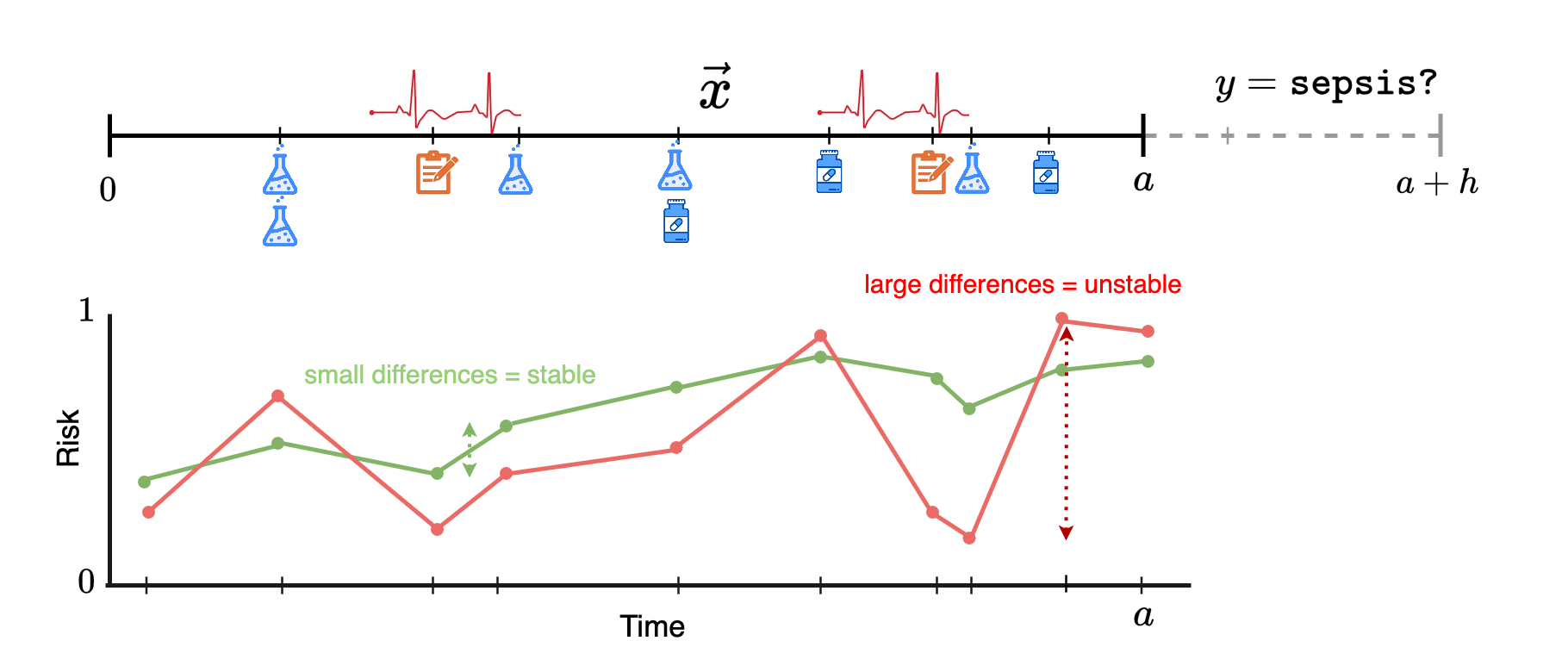} \label{fig:data-overview} 
    \caption{Example medical record of a patient and corresponding risk scores produced by a model at different timestamps. \textit{Top:} The input $x$ is $a$ hours of sparse time-series data consisting of tabular (blue), waveform (red), and text (orange) modalities, and output $y$ is a binary indicator of sepsis onset in the next $h$ hours. 
    \textit{Bottom:} Risk curves over time for patient $x$, for two models, shown in green (ideal) and red (inferior). Intuitively, the green predictions are more stable than red since the gaps between nearby points are smaller.
    Our stability measure quantifies the degree of fluctuations over time as the average local Lipschitz constant in short windows of time.
    } 
    \label{fig:overview}
\end{figure}

\textbf{Why temporal stability matters.} We argue that, in addition to discriminative performance, the \emph{temporal stability of risk scores} is critical for deployment. Clinicians consume risk trajectories to make intervention decisions; abrupt oscillations without new clinical evidence undermine interpretability, trust, and downstream workflow, contributing to alarm fatigue \cite{Butler_2025}. Standard AUROC and AUPRC metrics, especially when computed per-window, do not reflect temporal behavior at the per-patient level. Consequently, models with similar AUROC can produce markedly different trajectories—from plausible, evidence-aligned evolution to erratic fluctuations (Fig.~\ref{fig:overview}). A rigorous evaluation should therefore assess temporal stability alongside accuracy.
While some EEP models enforce smoother trajectories via loss terms \cite{yeche2023smoothing, yeche2024dsa}, stability has not been systematically evaluated.
This motivates a benchmark that measures stability alongside accuracy across clinical tasks.

To address this gap, we introduce \textbf{CAREBench}, a public benchmark for clinical EEP from multi-modal signals that emphasizes both predictive accuracy and temporal stability. CAREBench comprises six tasks spanning three open-source datasets--MC-MED, MIMIC-IV, and EHRShot--covering EHR, waveform, and text modalities (Table~\ref{table:hyperparameters}). Tasks include hypoglycemia and hyperkalemia (EHRShot), decompensation and sepsis (MC-MED), and ICU transfer and in-hospital mortality (MIMIC-IV). 

\textbf{Measuring temporal stability.} We quantify the local variability of a model's risk trajectory via a modified local Lipschitz constant. For a model $f$ and patient $x$, the stability score for a local window $c$ is:
\[L_c = \frac{1}{\left|T_c \right|}\sum_{t, t' \in T_c} \frac{|f(x_{\leq t}) - f(x_{\leq {t'}})|}{|t - t'|}
\]
where $x_{\leq t}$ is the data available up to time $t$ and $T_c$ is the set of timestamp pairs within $c$ hours, and $c$ is chosen based on expert expectations of clinically reasonable smoothness. 

\textbf{Models and evaluation.} We evaluate classical learners (XGBoost, Random Forests), deep sequence models (Transformers, Mamba), and zero-shot LLMs (Qwen3-32B) using standardized context lengths $a$ and horizons $h$ per task (Table~\ref{table:hyperparameters}). Classical models take as inputs derived statistics over tabular features,  neural models operate on raw sequences, and LLMs are evaluated zero-shot via structured prompting over EHR and text without any fine-tuning. 
Alongside stability, we report AUROC, AUPRC, F1, and a \emph{flip} count that measures the frequency of 0$\to$1 and 1$\to$0 transitions within the prediction window.


This work makes three main contributions:

\begin{itemize}[leftmargin=1.5em] 
    \item We introduce \textbf{CAREBench}, a multi-modal benchmark for clinical event prediction spanning six tasks across three public datasets.
    \item We formalize \textbf{temporal stability of risk scores} as a first-class evaluation dimension, proposing a simple, quantitative measure based on local Lipschitz constant.
    \item We provide a comprehensive evaluation of classical models, deep sequence models, and zero-shot LLMs on CAREBench, highlighting pronounced trade-offs between discriminative accuracy and temporal stability.
\end{itemize}

\section{Early Event Prediction with Multi-Modal Data}
\label{sec:method}

This section formalizes the prediction problem, defines how we construct evaluation instances, and introduces metrics for both accuracy and temporal stability.

\subsection{Problem Setting}
\label{subsec:problem}

We consider multiple data modalities $\mathcal{M}$, each with its own value domain $\mathcal{D}_m$. For example, ECG waveforms have $\mathcal{D}_{\text{wave}}=\mathbb{R}$ per sample, laboratory tests have $\mathcal{D}_{\text{lab}}=\mathcal{L}\times\mathbb{R}$ (a test identifier paired with a numeric value), and clinical notes have $\mathcal{D}_{\text{text}}$ of raw strings. A patient episode unfolds as a time-ordered sequence of observations:
\[
x_{[1:N]} = (o_1,\ldots,o_N), \qquad o_i=(m_i, d_i, t_i),
\]
where each observation $o_i$ specifies its modality $m_i\in\mathcal{M}$, data value $d_i\in\mathcal{D}_{m_i}$, and timestamp $t_i$. Timestamps are non-decreasing: $0 \le t_i \le t_{i+1}$, and at any reference time $T$, a model has access only to observations up to that point:
$x_{\le T} = \{\,o_i\in x_{[1:N]} : t_i \le T\,\}$.
We consider a specific adverse clinical event $E$ (e.g., sepsis onset, cardiac arrest) that serves as the prediction target.

\subsection{Task Construction}
\label{subsec:task}

An early event prediction model is a binary classifier trained to predict whether $E$ occurs within a fixed forecasting horizon $h > 0$. If $E$ occurs during an episode, let $t_E$ denote its first occurrence; otherwise the episode is labeled negative. At any reference time $T$, the episode is labeled positive if $E$ occurs within the next $h$ hours:
\[
y(T) \;=\; 
\begin{cases}
1 & \text{if } E \text{ occurs in } (T,\,T+h], \\
0 & \text{otherwise}.
\end{cases}
\]
Once trained, the model can be queried at any test-time reference point $T$, producing a risk score $f(x_{\le T})\in[0,1]$ that estimates $P(y(T)=1 \mid x_{\le T})$. 

Each patient episode contributes exactly one prediction instance to the evaluation set. We sample a single reference time $T$ per episode, subject to the constraints above. Importantly, to prevent models from learning spurious associations—such as ``longer observation histories indicate healthier patients''—we match the distribution of available history between positive and negative cases. Specifically, we use propensity-based sampling to ensure that the time since admission $T - t_1$ has a similar distribution across both outcome classes. 

\subsection{Evaluation Metrics}
\label{subsec:metrics}

\textbf{Accuracy.}
We report standard discrimination metrics on held-out data: AUROC, AUPRC, and F1. Given the rarity of adverse events, AUPRC is typically the most informative single metric. Unless stated otherwise, F1 is computed by thresholding the risk score at $\tau=0.5$.

\textbf{Temporal stability.}
Beyond point-wise accuracy, we assess whether risk trajectories evolve smoothly. We evaluate stability in a symmetric time window around $T$, defined by a probe radius $b>0$: specifically, $[T-b,\,T+b]$.  We quantify short-term variability using a \emph{local pairing window} of $c=10$ minutes, chosen in consultation with clinical experts to reflect a timescale over which abrupt oscillations would be clinically implausible. The distinction between $b$ and $c$ is important: $b$ defines \emph{where} we measure stability (around the prediction time), while $c$ defines \emph{how close} two timestamps must be to form a local comparison within that region. Hyperparameters $b$ and $h$ are task-specific and listed in Table~\ref{table:hyperparameters}.

We quantify the local variability of a model's risk trajectory via a modified local Lipschitz constant. For a model $f$ and patient history $x$, the stability score for a local window $c$ is:
\[
L_c \;=\; \frac{1}{|T_c|} \sum_{t, t' \in T_c} 
\frac{|f(x_t) - f(x_{t'})|}{|t - t'|},
\]
where $T_c$ is the set of local timestamp pairs $(t, t')$ and $t' - t \le c$ hours, within the evaluation window $[T-b, T+b]$.  
Lower values of $L_c$ indicate more stable trajectories; higher values suggest rapid, potentially erratic oscillations that undermine clinical trust.

\textbf{Alert volatility.}
As a complementary measure to stability, we count the number of times the risk score crosses the alert threshold $\tau$ (default 0.5) within the evaluation window $[T-b,\,T+b]$. We denote the binary alert state as $s(t)=1$ if $f(x_{\le t})\ge\tau$ and $s(t)=0$ otherwise. A \emph{flip} occurs whenever $s(t)$ changes between consecutive evaluated timestamps. We report the average flip count per patient. Higher flip counts indicate erratic alert behavior of repeatedly turning on and off, which is undesirable in continuous monitoring settings.

\section{CAREBench}

\subsection{Dataset Description}
\label{sec:dataset}

We use three public datasets: EHRShot \cite{wornow2023ehrshot}, Multimodal Clinical Monitoring in the Emergency Department (MC-MED) \cite{mc-med}, and Medical Information Mart for Intensive Care (MIMIC)-IV \cite{mimic-iv}. 

\textbf{MC-MED}, released in March 2025, represents a significant advancement in publicly available medical time-series datasets.
It includes 118K emergency department (ED) visits between 2020 and 2022 from 70K unique patients, and provides both high-frequency waveform and traditional tabular data.
MC-MED integrates continuously monitored vital signs and physiologic waveforms (ECG, photoplethysmogram, respiration) with clinical events and outcomes. 
The availability of long waveform segments at this scale, paired with EHR data, enables the analysis of rapidly evolving patient states and intervention responses at unprecedented temporal resolution.

\textbf{MIMIC-IV} is one of the most widely used large-scale medical datasets, encompassing both ED and ICU visits. 
Released in August 2019, MIMIC-IV contains records from over 65K ICU visits and 200K emergency department visits at Beth Israel Deaconess Medical Center. 
The data comprises diverse sources, including vital signs, laboratory tests, radiology reports, and clinical notes. 
MIMIC-IV contains the complete patient trajectory from emergency department admission to ICU discharge, providing new opportunities to model and predict patient evolution over time.

\textbf{EHRShot} provides de-identified structured data from 6,739 patients from Stanford Medicine. 
Although smaller in size and uni-modal compared to the other two datasets, EHRShot complements them by providing longitudinal records spanning several years that are not restricted to ICU or ED settings.
Furthermore, as EHRShot was collected to assess the few-shot capabilities of LLMs, it includes 15 prediction tasks across diverse clinical settings, along with a framework for evaluating not only accuracy but also sample efficiency and task adaptation.

\subsection{Task Description}
We curate a total of six tasks as summarized in Table \ref{table:hyperparameters}. Specifically, we select hypoglycemia and hyperkalemia from EHRShot, sepsis and decompensation \cite{mc-bec} from MC-MED, and ICU transfer and in-hospital mortality from MIMIC-IV.

\begin{table}[t]
\footnotesize
    \centering
    \caption{Configurations for each dataset in \textbf{CAREBench}.}
    \vspace{3mm}
    \begin{tabular}{l|cccccccc}
        \textbf{Task} & Dataset & Modality & Size & Prevalence (\%) & $h$ (hours) & $b$ (hours) \\
                \toprule
        Hyperkalemia & EHRShot & EHR & 5.9K & 4.04 &  1  & 0.5, 4.0  \\
        Hypoglycemia & EHRShot & EHR & 5.9K & 5.85 &  1  & 0.5, 4.0  \\
        Decompensation & MC-MED & EHR, Waveform & 68K & 13.2 & 1.5 & 0.5, 4.0  \\
        Sepsis & MC-MED & EHR, Waveform & 11K & 3.63 & 1.5 & 0.5, 4.0 \\
        ICU Transfer & MIMIC-IV & EHR & 60K & 8.98 & 6 & 3.0   \\  
        Mortality & MIMIC-IV & EHR, Text & 30K & 12.54 & 12 & 3.0 \\
    \end{tabular}\label{table:hyperparameters}
\end{table}




\textbf{Hyperkalemia \& Hypoglycemia.} 
In EHRShot, hyperkalemia and hypoglycemia are originally framed as four-way classification tasks, predicting whether lab results will be normal, mild, moderate, or severe \textit{within the next minute}. We adapt these tasks to a longer horizon of $h = 1$ hour and reformulate them as binary classification problems, predicting whether the lab result will be severe.
For these tasks, a one- to two-hour warning is clinically valuable, providing a critical window for preventive interventions, such as administering glucose for hypoglycemia \cite{cryer2007hypoglycemia} or initiating rapid-acting therapies for hyperkalemia \cite{kovesdy2015hyperkalemia}.

Concretely, for hyperkalemia, we follow EHRShot and define an event as true if the potassium level exceeds 7 mmol/L, considering all lab results coded as \texttt{LOINC/LG7931-1}, \texttt{LOINC/LP386618-5}, \texttt{LOINC/LG10990-6}, \texttt{LOINC/6298-4}, or \texttt{LOINC/2823-3}. 
For hypoglycemia, an event is defined as true if the glucose level is below 3 mmol/L, considering lab results coded as \texttt{SNOMED/33747003}, \texttt{LOINC/LP416145-3}, or \texttt{LOINC/14749-6}. 

\textbf{Decompensation.} 
In the MC-MED dataset, decompensation (i.e., clinical deterioration) is defined as follows: given the first 15 minutes of data since admission, predict whether the patient will develop new-onset tachycardia, hypotension, or hypoxia within the next 30, 60, or 90 minutes \cite{mc-bec}. 
We use the same labels but set the forecasting horizon to $h = 1.5$ hours, allowing the input length to vary for each patient.
To prevent trivial predictions, the cohort is restricted to patients presenting with initially normal vital signs.

\textbf{Sepsis.} 
EEP for sepsis has been widely studied due to its clinical importance \cite{shashikumar2021sepsis, moor2023multicenter, moor2021sepsis}. Sepsis carries a high mortality rate, accounting for one in three ICU deaths in the United States \cite{singer2016sepsis}. 
Studies have shown that each hour of delayed treatment is associated with a 4--9\% increase in mortality \cite{time-to-treatment,outcome-trews,ews-monitoring}.
However, clinical deployments have reported mixed results regarding the usefulness of such models \cite{shashikumar2025sepsis}. In particular, many commercial models \cite{wong2021epic, henry2022trews} perform well in validation but often fail to generalize to real-world settings.

More importantly, these models have been largely limited to EHR data, primarily because publicly available datasets such as MIMIC-III \cite{mimic3-benchmarks} and PhysioNet 2019 \cite{reyna2020physionet} are tabular. 
In contrast, we use MC-MED, which includes which includes detailed EHR data, such as ventilator settings, medications, and minute-level vitals, alongside high-resolution waveform data.
For sepsis onset prediction, we first narrow the cohort to patients at risk, following prior work \cite{ginestra2019clinician}, and compute labels according to the Sepsis-3 definition \cite{singer2016sepsis}. 
See Appendix~\ref{appendix:definition} for further details on the cohort definition and eSOFA criteria.
We set the prediction horizon to $h = 1.5$ hours, as the majority of positive cases in this dataset meet the criteria within two hours of admission.

\textbf{ICU Transfer.} 
This task involves predicting whether a patient admitted to the ED will be transferred to the ICU.
We collect the data from MIMIC-IV-ED and set the prediction horizon to $h = 6$ hours.
The available ED data include seven vital signs (temperature, heart rate, respiratory rate, oxygen saturation, systolic and diastolic blood pressure, and acuity), as well as cardiac rhythm alarms and medications. The median number of vital readings per patient is four, with a maximum of 110.

\textbf{In-hospital mortality.} 
In-hospital mortality prediction using the MIMIC-IV dataset has been extensively studied \cite{Pang_2022, Sun_2023}, including recent work employing large language models (LLMs) \cite{lovon20204mimicllm}. Accordingly, we include this task for completeness.
For mortality prediction, input data and labels are derived from ICU stay records in MIMIC-IV, including medications, labs, and vital signs, with a prediction horizon of $h = 12$ hours.
When prompting the LLM, we also include the most recent radiology report, if available, from MIMIC-IV-Notes.

\section{Baselines}
\label{sec:baselines}

We evaluate six models, ranging from classical machine learning methods to LLMs.

\textbf{Classical models.}
We include XGBoost and Random Forest, both of which have demonstrated strong performance in clinical prediction tasks \cite{burger2024foundationmodelscriticalcare}.
For these models, we compute summary statistics for the 30 most frequent numeric features, excluding text. Features with more than 90\% missing values are removed, and remaining missing values are imputed using the feature-wise median.
We use the model logits directly as risk scores.

\textbf{Deep learning models.}
We evaluate Mamba-130M \cite{mamba} and GPT-2 Small (124M) \cite{gpt2}.
Mamba is a state-space model developed to effectively model long-range dependencies.
We use small models due to limited dataset availability and computational constraints. 
Following prior work demonstrating that pretraining improves performance \cite{wornow2025contextcluesevaluatinglong},
we pretrain the models with next-token prediction objectives before supervised fine-tuning.
During training, we create a custom tokenizer for each dataset to o handle the non-standardized and highly variable EHR codes across hospital sites.

\textbf{Auto-regressive modeling.}
We also evaluate GPT-2 in an auto-regressive forecasting setup. 
Unlike standard classification approach, where the final linear layer receives logits directly, the auto-regressive approach aggregates logits from each unrolled prediction step. 
Given an input sequence $x_{[1:N]}$ and a forecasting horizon $h$, the model iteratively predicts $x_{[N+1:N+h]}$. Specifically, $x_{[N+1]} = (m_{N+1}, d_{N+1}, t_{N+1})$ is predicted from $x_{[1:N]}$ and appended to form $x_{[1:N+1]}$, which is then used to predict $x_{[N+2]}$, and so on until $x_{[N+h]}$ is obtained. 
The resulting predicted sequence $x_{[N+1:N+h]}$ is passed through a linear layer to generate the final risk score.

\textbf{LLMs.}
Due to data usage restrictions, we assess the zero-shot performance of LLMs using \texttt{Qwen3-32B} \cite{alibaba2015qwen3}, the leading open-source model on the medical QA benchmark MEDIC \cite{kanithi2024medic}.
We use a chain-of-thought prompting with detailed instructions derived from clinical case studies \cite{heil2020Sepsis} (see Appendix~\ref{sec:appendix-prompts}).
We use the model’s verbalized risk estimates. While early studies reported poor calibration of verbalized confidence scores \cite{xiong2024llm}, later work demonstrated that structured prompting templates can yield well-calibrated outputs, particularly for models larger than 70B parameters \cite{yang2024verbalizedconfidencescoresllms}.
We limit the total sequence length to 16K tokens, including the system prompt, and left-truncate longer sequences to retain the most recent information.

\section{Evaluation}
\label{sec:evaluation}

\subsection{Experimental Setup}
\textbf{Data pre-processing.} 
After pre-processing to align the distributions of time since admission (\S \ref{subsec:task}), we truncate the final $h$ hours of data from both positive and negative samples to account for the forecasting horizon.
This process removes positive samples where the first event onset $t_1$ occurs within the first $h$ hours since admission, corresponding to cases in which the onset is too soon for clinically meaningful forecasting.

\textbf{Validation \& Testing.} 
We employ a five-fold cross-validation strategy for each task. Hyperparameter optimization is conducted on the first two folds, consistent with the methodology outlined in \cite{burger2024foundationmodelscriticalcare}. 
In contrast, due to our zero-shot evaluation of LLMs, we directly apply them to a single fold of the test set without any training.
Further details on compute resources and training hyperparameters are provided in Appendix \ref{sec:appendix-hyperparameter}.

\subsection{Performance}

\begin{table}[t]
\centering
\caption{Performance on EHRShot benchmarks hyperkalemia and hypoglycemia.}
\vspace{2mm}
\footnotesize
\label{table:ehrshot-short}
\begin{tabular}{c|cccc|cccc}
    \toprule
    & \multicolumn{4}{c|}{\textbf{Hyperkalemia}}  & \multicolumn{4}{c}{\textbf{Hypoglycemia}} \\
    \midrule
    \textbf{Method} & AUROC & AUPRC & F1 & Stability & AUROC & AUPRC & F1 & Stability \\
    \midrule
    XGB  & {0.6244} & 0.0959 & {0.1499} & 0.0183  & 0.7470  & 0.2371  & \textbf{0.3175} & 0.0183  \\
    RF   & 0.6459 & 0.1078  & 0.0896 & 0.0092 & 0.7759 & \textbf{0.2710} & 0.2719 & 0.0092 \\
    Mamba & 0.7748 & 0.1552 & 0.1352 & 0.0171 & 0.7454 & 0.2017 & 0.2190 & 0.0069 \\
    GPT2  & \textbf{0.8116} & \textbf{0.1561} & \textbf{0.1746} & 0.0222 & 0.7341 & 0.1808 & 0.2118 & 0.0273 \\
    GPT2$_{\text{AR}}$ & 0.7838 & 0.1471 & 0.1497 & \textbf{0.0000} & \textbf{0.7919} & 0.1585 & 0.2515 & \textbf{0.00001} \\
    Qwen3-32B & 0.7171 & 0.0771 & 0.1365 & 10.6661 & 0.7726 & 0.1185 & 0.1860 & 11.6781 \\
    \bottomrule
\end{tabular}
\end{table}


\begin{table}[t]
\centering
\footnotesize
\caption{Performance on MC-MED benchmarks decompensation and sepsis. }
\vspace{2mm}
\label{table:mcmed-short}
\begin{tabular}{c|cccc|cccc}
    \toprule
     & \multicolumn{4}{c|}{\textbf{Decompensation}} & \multicolumn{4}{c}{\textbf{Sepsis}} \\
    \midrule
    \textbf{Method} & AUROC & AUPRC & F1 & Stability & AUROC & AUPRC & F1  & Stability \\
        \midrule
    XGB     & 0.6855 & 0.2462  & 0.3208 & 0.0452 & 0.6893 & 0.0893 & \textbf{0.1190} & 0.0494 \\
    RF      & 0.6765 & 0.2364 & 0.2931 & 0.0258 &  \textbf{0.6935} & 0.0866 & 0.0612 & 0.0202 \\
    Mamba   & 0.6962 & 0.2600 &0.3219 & 0.0795 & 0.6143 & 0.0683 & 0.0863 & 0.0420\\
    	
    GPT2    & \textbf{0.6978} & \textbf{0.2900} & \textbf{0.3235} & \textbf{0.0181} &  0.6564 & 0.0670 & 0.0970 & 0.0150 \\
    GPT2$_{\text{AR}}$ & 0.6514 & 0.2418 & 0.2670 & 0.1025  & 0.6583 & \textbf{0.0935} & 0.0659 & \textbf{0.0001} \\
    Qwen3-32B & 0.5508 & 0.1496 & 0.2332 & 0.1559 & 0.6154  & 0.0494 & 0.1017 &  0.1375 \\ 
    \bottomrule
\end{tabular}
\vspace{-2mm}
\end{table}

To compare predictive accuracy, we report AUROC, AUPRC, and F1 for all six tasks in Tables \ref{table:ehrshot-short}, \ref{table:mcmed-short}, and \ref{table:mimic-short}, and provide the full result including the flip counts and 1-sigma standard deviation in Appendix \ref{appendix:full-results}. 
Overall, the results show no clear winner.
Similar to prior work \cite{burger2024foundationmodelscriticalcare}, the ensemble models XGB and RF perform well across tasks, achieving AUROC within 0.0492 of the best performer, except for 
\begin{table}[H]
\centering
\footnotesize
\caption{Performance on MIMIC-IV benchmarks in ICU transfer and in-hospital mortality.}
\vspace{2mm}
\label{table:mimic-short}
\begin{tabular}{c|cccc|cccc}
    \toprule
     & \multicolumn{4}{c|}{\textbf{ICU Transfer}} & \multicolumn{4}{c}{\textbf{In-hospital Mortality}} \\
    \midrule
    \textbf{Method} & AUROC & AUPRC & F1 & Stability & AUROC &  AUPRC & F1 & Stability \\
        \midrule
    XGB & 0.8869 & 0.7868  & 0.6209 & 0.0006 & 0.8116 & 0.3282 & 0.3000 & 0.0104 \\
    RF & 0.8681 & \textbf{0.8364} & \textbf{0.6599} & 0.0191 & 0.8068 & 0.2818 & 0.0000 & 0.0197 \\
    Mamba & 0.8707 & 0.4834 & 0.3551 & 0.0621 & 0.7870 & 0.4311 & 0.3652 & 0.0276 \\
    GPT2 & \textbf{0.8875} & 0.5353 & 0.3648 & 0.0632 & 0.7600 & 0.3804 & 0.2874 & 0.0320 \\
    GPT2$_{\text{AR}}$ & 0.8598 & 0.4596 & 0.3678 & \textbf{0.0001} & 0.7107 & 0.3630 & 0.2886 & \textbf{0.0001} \\
    Qwen3-32B & 0.7515 & 0.5565 & 0.5587  & 1.3675 & \textbf{0.8608} & \textbf{0.5849} & \textbf{0.5786} & 0.3035 \\
    \bottomrule
\end{tabular}
\end{table}
hyperkalemia prediction.
This can be partly attributed to the dataset characteristics, as EHRShot is uni-modal and relatively small compared to the other two datasets.

For deep learning models, we observe comparable performance across the three approaches, with GPT-2 emerging as the overall best performer by a slight margin.
Notably, GPT-2 achieves the highest AUROC among all models for half of the tasks.
Although the auto-regressive variant GPT2$_{\text{AR}}$ attains the highest AUROC for hypoglycemia task, the improvement is marginal relative to the substantial computational overhead required to unroll predictions during training and testing.

The zero-shot evaluation of Qwen3-32B yields mixed results: it performs well on MIMIC-IV tasks in terms of both AUROC and AUPRC, but struggles with the other datasets. This discrepancy likely reflects the nature of the tasks, as in-hospital mortality and ICU transfer are general and rely on broad clinical concepts and domain knowledge. Moreover, these tasks include text data from radiology reports, which is naturally suited for LLMs. These findings suggest that further research is needed to effectively leverage LLMs for structured tabular and waveform data. The strong performance of Qwen3 on in-hospital mortality prediction, outperforming all other models despite receiving no task-specific training, highlights the potential of zero-shot LLM applications.

\subsection{Stability}
To evaluate stability, we freeze the model described in \S5.2 and vary the length of the input data by adding or removing measurements at the end of the sequence, generating multiple risk scores for the fixed prediction horizon $h$.
Overall, classical models exhibit stable predictions, primarily because their inputs consist of derived statistics rather than raw data. 
Limiting the input to 30 features further enhances stability, as each new measurement has a limited impact compared to deep learning models and LLMs, where every variation introduces additional tokens and produces new predictions.
 
\begin{figure}[t]
\centering
\begin{tikzpicture}
\begin{axis}[
    xlabel={Hours},
    width=10cm,   
    height=5cm, 
    ylabel={Risk Score},
     xmin=0,         
    ymax=1.0, 
    xmax=3.8,
    legend style={at={(1.0,1)}, anchor=north west},
    grid=both
]



\addplot[color=olive, ultra thick] coordinates {
    (0.5, 0.35015538)
    (0.6, 0.15374681)
    (0.7, 0.07814392)
    (0.8, 0.45238423)
    (0.9, 0.1978732)
    (1.0, 0.31805396)
    (1.1, 0.82485783)
    (1.2, 0.838816)
    (1.3, 0.85960793)
    (1.4, 0.856828)
    (1.5, 0.8666854)
    (1.6, 0.89230734)
    (1.7, 0.93627024)
    (1.8, 0.93279505)
    (1.9, 0.9564865)
    (2.0, 0.96429485)
    (2.1, 0.9780293)
    (2.2, 0.9831335)
    (2.3, 0.98492765)
    (2.4, 0.9877101)
    (2.5, 0.9929831)
    (2.6, 0.9982597)
    (2.7, 0.99806637)
    (2.8, 0.99816304)
    (2.9, 0.99943453)
    (3.0, 0.9994734)
    (3.1, 0.9998104)
    (3.2, 0.99993634)
    (3.3, 0.999972)
    (3.4, 0.9999807)
    (3.5, 0.99997807)
    (3.6, 0.9999919)
    (3.7, 0.9999907)
    (3.8, 0.9999856)
};
\addlegendentry{Mamba}

\addplot[color=teal, ultra thick] coordinates {
    (0.5,0.35662198)
    (0.6,0.3538254)
    (0.7,0.35260603)
    (0.8,0.36023507)
    (0.9,0.35434166)
    (1.0,0.36632648)
    (1.1,0.38873893)
    (1.2,0.37066063)
    (1.3,0.38606638)
    (1.4,0.368343)
    (1.5,0.3763004)
    (1.6,0.39118132)
    (1.7,0.39752457)
    (1.8,0.38697815)
    (1.9,0.38380483)
    (2.0,0.39863306)
    (2.1,0.39495403)
    (2.2,0.40065452)
    (2.3,0.39899886)
    (2.4,0.41167113)
    (2.5,0.42708796)
    (2.6,0.4254886)
    (2.7,0.42529798)
    (2.8,0.4296824)
    (2.9,0.4191288)
    (3.0,0.43885684)
    (3.1,0.4375471)
    (3.2,0.4545566)
    (3.3,0.44790104)
    (3.4,0.43896928)
    (3.5,0.44468814)
    (3.6,0.4472369)
    (3.7,0.45959324)
    (3.8,0.46113193)
};

\addlegendentry{GPT2}

\addplot[color=magenta, ultra thick] coordinates {
(0.5, 0.4995759)
(0.6, 0.49957162)
(0.7, 0.49959913)
(0.8, 0.49961466)
(0.9, 0.49964085)
(1.0, 0.49966952)
(1.1, 0.49965703)
(1.2, 0.499658)
(1.3, 0.4997143)
(1.4, 0.4997147)
(1.5, 0.49976766)
(1.6, 0.49977526)
(1.7, 0.49978295)
(1.8, 0.4998092)
(1.9, 0.49986675)
(2.0, 0.4999144)
(2.1, 0.49996066)
(2.2, 0.4999951)
(2.3, 0.49994618)
(2.4, 0.50007963)
(2.5, 0.5000706)
(2.6, 0.50014514)
(2.7, 0.5001933)
(2.8, 0.50022686)
(2.9, 0.500276)
(3.0, 0.50035614)
(3.1, 0.50035644)
(3.2, 0.50053215)
(3.3, 0.50048095)
(3.4, 0.50054437)
(3.5, 0.5005701)
(3.6, 0.50072914)
(3.7, 0.50072205)
(3.8, 0.50087214)
};
\addlegendentry{GPT2$_\text{AR}$}

\addplot[color=violet, ultra thick] coordinates {
    (0.000, 0.75)
    (0.033, 0.85)
    (0.150, 0.85)
    (0.183, 0.7)
    (0.217, 0.65)
    (0.317, 0.3)
    (0.350, 0.85)
    (0.383, 0.85)
    (0.400, 0.85)
    (0.483, 0.85)
    (0.517, 0.75)
    (0.550, 0.85)
    (0.650, 0.75)
    (0.683, 0.85)
    (0.767, 0.85)
    (0.800, 0.85)
    (0.817, 0.85)
    (0.833, 0.85)
    (0.850, 0.85)
    (0.867, 0.85)
    (0.883, 0.85)
    (0.900, 0.85)
    (0.933, 0.65)
    (1.000, 0.25)
    (1.100, 0.85)
    (1.133, 0.35)
    (1.167, 0.3)
    (1.267, 0.7)
    (1.300, 0.75)
    (1.333, 0.75)
    (1.433, 0.85)
    (1.467, 0.85)
    (1.500, 0.85)
    (1.600, 0.9)
    (1.633, 0.95)
    (1.667, 0.95)
    (1.767, 0.95)
    (1.800, 0.95)
    (1.817, 0.9)
    (1.900, 0.95)
    (1.917, 0.95)
    (1.933, 0.95)
};
\addlegendentry{Qwen3-32B}

\end{axis}
\end{tikzpicture}
\caption{Risk score of a patient over the 4 hour horizon for sepsis onset prediction, plotted for Mamba, GPT-2, GPT2$_\text{AR}$, and Qwen3-32B models. The true label is \texttt{False}.}
\label{fig:stability}
\end{figure}
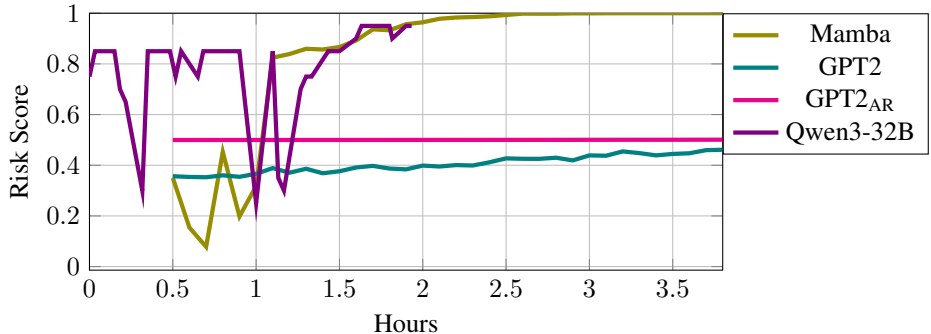

Unlike classical models, deep learning models—and particularly LLMs—exhibit greater variability in risk score predictions over time, as shown in Figure \ref{fig:stability}.
Qwen3 consistently shows the lowest stability across all tasks, with stability scores an order of magnitude higher than those of other models. For example, the stability scores are 10.67 for hyperkalemia and 11.68 for hypoglycemia, compared to less than 0.03 for other models.

On the other hand, GPT2$_{\text{AR}}$ shows an interesting trade-off: while it struggles with longer forecasting horizons, such as 6 and 12 hours for ICU transfer and in-hospital mortality predictions, it achieves markedly better stability, with near-zero Lipschitz constants for hyperkalemia and hypoglycemia tasks.
This is supported by the flip counts reported in Appendix \ref{appendix:full-results}, where GPT2$_{\text{AR}}$ has far fewer prediction flips—for example, 0.31 for hyperkalemia compared to 1.71 for GPT-2.

\section{Limitations and Future Work}
\label{sec:limitations}

\textit{Metrics.} 
CAREBench focuses on two key aspects: accuracy under class imbalance and stability. However, several other factors are critical for clinical adoption, including calibration of risk scores (particularly for LLMs), interpretability of predictions, and stability of explanations produced by post-hoc explainability methods \cite{ghorbani2018interpretationneuralnetworksfragile}. A clinically useful future direction is to conduct a multi-center study to evaluate how models generalize under distribution shifts across hospital sites.

Furthermore, an important future direction is refining the stability score to better reflect clinical utility. 
The current metric does not account for underlying changes in a patient’s condition, which can result in substantial increases or decreases in actual risk. 
Additionally, as observed in our experiments, the score is influenced by measurement sparsity, with stability scores typically worse on dense datasets such as MC-MED compared to EHRShot. 
A refined metric should normalize stability according to both the patient’s clinical condition and the characteristics of the dataset.

\textit{Task Types}. 
As CAREBench is designed for EEP using time-series clinical data, many standard prediction tasks, such as readmission and length of ICU stay \cite{wornow2023ehrshot}, where forecasting horizon spans months or years, are out of scope.
While the six tasks included in the benchmark are carefully selected to cover diverse clinical settings from ED to ICU, they are limited to binary predictions, excluding regression-based or multi-event tasks. 

\textit{Models.} 
While our evaluation aims to encompass model families ranging from classical approaches to LLMs, several directions remain for future work, particularly with respect to LLMs. 
These include supervised fine-tuning, Reinforcement Learning with Verifiable Rewards (RLVR) \cite{deepseek}, advanced prompting techniques such as tree-of-thought reasoning and majority voting, and the exploration of agentic LLMs \cite{mehandru2024agents}.

While Qwen3 is the leading open-source model on the MEDIC benchmark, medical LLMs finetuned on clinical text may offer better domain alignment.
However, models such as MeLLaMA \cite{xie2024llama}, which is fine-tuned from LLaMA2, have limited context windows of 4096 tokens and therefore cannot accommodate the large prompts required for our evaluation, where the median test prompt size for MIMIC-IV is 15K tokens. 
A promising direction for future work is to develop prompt compression methods, either by filtering nonessential lab measurements or by designing more advanced tokenization techniques.

\section{Related Work}

\textbf{Medical Datasets.} 
In addition to the datasets outlined in \S \ref{sec:dataset}, three other major datasets have significantly influenced healthcare time-series benchmarks: eICU \cite{pollard2018eicu}, which includes over 200,000 ICU admissions with diverse clinical data; 
AmsterdamUMCdb \cite{thoral2021amsterdam}, containing 1 billion clinical data points from 23,106 ICU admissions in the Netherlands; 
and HiRID \cite{eth_zurich_hirid_49}, comprising 34,000 admissions with high-resolution two-minute interval monitoring.

\textbf{Existing Benchmarks.}
Prior benchmarks for healthcare time-series prediction have primarily focused on accuracy, with limited attention to temporal smoothness, a key factor for clinical utility.
Yet Another ICU Benchmark (YAIB) \cite{vandewater2024yaib} offers a modular framework for reproducible model evaluation across public datasets, including MIMIC, eICU, HiRID, and AmsterdamUMCdb. 
YAIB benchmarks on five prediction tasks and highlights the impact of data curation and preprocessing choices on model performance. 
While comprehensive in its evaluation framework, YAIB does not explicitly incorporate smoothness metrics, which our work introduces.

Similarly, Burger et al. \cite{burger2024foundationmodelscriticalcare} developed the largest harmonized dataset for time-series evaluation, comprising 600K visits and 1 billion data points across six tasks and seven EHR datasets, including MIMIC-IV, eICU, and HiRID. 
Other multi-center studies \cite{moor2023multicenter} have facilitated cross-dataset benchmarking, assessing how model performance generalizes across diverse clinical settings. 
However, these studies primarily focus on predictive accuracy under distribution shifts and do not explicitly evaluate temporal prediction stability.

FoMoH \cite{fomoh} benchmarks six foundation models against supervised classical models. While FoMoH emphasizes calibration, fairness, and computational cost in addition to predictive accuracy, CAREBench highlights temporal stability as an additional critical quality for deployment. Unlike CAREBench, FoMoH is limited to the tabular modality. In contrast, {MIMIC-III-Ext} \cite{yin2025mimic-ext} is a curated subset of MIMIC-III designed for irregular multivariate clinical time-series forecasting. It addresses challenges such as missing data and variable sampling rates, focusing on next-value prediction rather than the risk of specific adverse clinical events.

\textbf{Models \& Training.} 
The unique challenges of medical time-series data have necessitated specialized approaches for model training. 
Prior work has shown that the irregularity in the positional nature of EHR data, compared to text, makes next-token prediction more difficult \cite{wornow2025contextcluesevaluatinglong}, motivating the development of custom architectures based on transformers \cite{song2024trajgpt}, Mamba \cite{wornow2025contextcluesevaluatinglong}, and graph neural networks \cite{zhang2024gnnmedical}. 
Many studies also explore custom tokenization techniques that explicitly encode the time delta between EHR records \cite{ethos,cherxgpt}. 
These architectures, however, typically focus exclusively on the tabular modality.
Moreover, although many LLM-based and foundation models have demonstrated promising performance, they have primarily been applied to univariate time-series data, such as ECG or vital signs \cite{chan2024medtsllmleveragingllmsmultimodal, ehr2path}, or to data without temporal aspects \cite{nep, transformerEHR}. Prior works have also proposed training objectives inspired by adversarial learning \cite{namadversarial2024} to improve representations of clinical time-series data.

\section{Conclusion}
In this paper, we introduced \textbf{CAREBench}, a novel benchmark for medical early event prediction that uniquely emphasizes prediction stability alongside accuracy across six diverse tasks on three public datasets. Our evaluation of classical, deep learning, and LLM-based approaches reveal a critical trade-off between accuracy and stability, with no single model excelling across all metrics. While LLMs demonstrate promising accuracy, they exhibit poor stability, whereas auto-regressive approaches achieve excellent stability, often at the cost of accuracy. 
CAREBench addresses an important gap in existing evaluations by quantifying prediction stability, a crucial factor for clinical adoption, and provides a foundation for developing models that balance performance with the temporal consistency required for real-world healthcare implementation.

\begin{ack}
This research was supported by ARPA-H program on Safe and Explainable AI under the award D24AC00253-00.
\end{ack}

\bibliography{references}


\newpage
\appendix
\section{Sepsis Cohort Definition and eSOFA Criteria} \label{appendix:definition}

For sepsis prediction, the cohort includes patients suspected of having sepsis, and the label is determined based on the eSOFA criteria.

In particular, the inclusion criteria is:
\begin{itemize}
    \item Admission source of ED
    \item Temperature of $< 36^\circ\text{C}$ or $> 38.5^\circ\text{C}$ within 24 hours of admission (\texttt{Temp\_time})
    \item At least one of the following within 24 hours of admission:
        \begin{itemize}
            \item WBC Count $> 12K$ or $<4K/\mu L$ (\texttt{WBC\_time})
            \item HR $> 90$ (\texttt{HR\_time})
            \item RR $> 20$ (\texttt{RR\_time})
        \end{itemize}
    \item At least 1 of the \texttt{WBC\_time}, \texttt{HR\_time}, \texttt{RR\_time} within 12 hours of \texttt{Temp\_time}
    \item No intravenous antibiotic at or before the time of the first criteria met,
\end{itemize}

and eSOFA criteria \cite{Rhee2017-eSOFA} is
\begin{itemize}
    \item Presumed serious infection 
    \begin{itemize}
        \item Blood culture obtained (regardless of the results),
        \item $\geq 4$ QADs starting within $\pm$ 2 days of \texttt{blood\_culture\_day}
    \end{itemize}
    \item any 1 of below within $\pm$ 2 days of \texttt{blood\_culture\_day} (acute organ dysfucntion)
    \begin{itemize}
        \item Vassopressor initiation
        \item Initiation of mechanical ventilation
        \item Doubling in serum creatinine level or decrease by $\geq 50\%$ of eGFR (excluding patients with end-stage kidney disease [585.6])
        \item Total bilirubin level $\geq$ 2.0mg/dL and doubling
        \item Platelet count $<$ 100 cells/$\mu$L and $\geq$ 50\% decline from baseline (excluding baseline $<$ 100 cells/$\mu$L)   \item Serum lactate $\geq$ 2.0 mmol/L
    \end{itemize}
    
\end{itemize}

To accommodate the short time frames in the emergency department, where patients typically stay for at most six hours, we relax the eSOFA criterion from requiring four days to 72 hours of consecutive antibiotic administration.

\section{Experimental Setup} \label{sec:appendix-hyperparameter}

We run evaluation
on a cluster with 10 NVIDIA A100 80GB GPUs and 96 Intel Xeon Gold CPU cores. 
The average training time is approximately 24 hours for 100 epochs per dataset.

\textbf{Hyperparameters.}
For XGBoost, we set the learning rate to 0.1 and train for 100 epochs using the following hyperparameters:
\texttt{max\_depth}=5, \texttt{min\_child\_weight}=1, \texttt{gamma}=0, and \texttt{subsample}=0.8.
For Random Forest, we use \texttt{n\_estimators}=100, \texttt{max\_depth}=10, and \texttt{min\_samples\_split}=2.
For deep learning models, we train for 100 epochs using the best learning rate selected from $\{1e^{-5},\, 5e^{-5},\, 1e^{-4}\}$.
For {Qwen3-32B}, we report results using the \texttt{no-thinking} mode and the full \texttt{unquantized} model, which achieved the best validation performance.
The decoding configuration is: $\mathtt{temperature}=0.6$, $\mathtt{top\text{-}k}=0.8$, $\mathtt{top\text{-}p}=20$, and $\mathtt{min\text{-}p}=0$.

\section{Full Experimental Results} \label{appendix:full-results}

\subsection{Main Result}
The full results of the baselines models on CAREBench tasks are reported in Tables \ref{table:hyperkalemia}--\ref{table:mcmed-sepsis}.
Standard deviations are not reported for Qwen3 because we perform zero-shot evaluation without any training.

\begin{table}[th]
\caption{Performance on hyperkalemia prediction with EHRShot dataset.}
\vspace{3mm}
\label{table:hyperkalemia}
\begin{adjustbox}{center,max width=\textwidth}
\begin{tabular}{c|ccc|ccc}
    \toprule
    \textbf{Method} & AUROC $\uparrow$ & AUPRC $\uparrow$ & F1 $\uparrow$ & Stability $\downarrow$ & Flip $\downarrow$ \\
    \midrule
    XGB   & \std{0.6244}{0.0314} & \std{0.0959}{0.0240} & \std{0.1499}{0.0671} & {0.0183} & 0.5949 \\ 
    RF    & \std{0.6459}{0.0511} & \std{0.1078}{0.0346} & \std{0.0896}{0.0469} & {0.0092}  & 0.6118 \\ 
    Mamba & \std{0.7748}{0.0232} & \std{0.1552}{0.0381} & \std{0.1352}{0.0181} & 0.0171 & 1.1772 \\ 
    GPT2  & \std{0.8116}{0.0099} & \std{0.1561}{0.0276} & \std{0.1746}{0.0083} & 0.0222 & 1.7089 \\ 
    GPT2$_{\text{AR}}$ & \std{0.7838}{0.0032} & \std{0.1471}{0.0386} & \std{0.1497}{0.0214} & 0.0010 & 0.3077 \\ 
    Qwen  & 0.7171 & 0.0771 & 0.1365 & 0.2819 & 2.8310 \\ 
    \bottomrule
\end{tabular}
\end{adjustbox}
\end{table}

\begin{table}[ht]
\caption{Performance on hypoglycemia prediction with EHRShot dataset.}
\label{table:hypoglycemia}
\vspace{3mm}
\begin{adjustbox}{center,max width=\textwidth}
\begin{tabular}{c|ccc|ccc}
    \toprule
    \textbf{Method} & AUROC $\uparrow$ & AUPRC $\uparrow$ & F1 $\uparrow$ & Stability $\downarrow$ & Flip $\downarrow$ \\
    \midrule
    XGB   & \std{0.7470}{0.0261} & \std{0.2371}{0.0300} & \std{0.3175}{0.0318} & {0.0153} & 0.5677 \\ 
    RF    & \std{0.7759}{0.0296} & \std{0.2710}{0.0223} & \std{0.2719}{0.0204} & {0.0069}  & 0.5966 \\ 
    Mamba & \std{0.7454}{0.0054} & \std{0.2017}{0.0149} & \std{0.2190}{0.0314} & 0.0229  & 1.9747 \\ 
    GPT2  & \std{0.7341}{0.0062} & \std{0.1808}{0.0107} & \std{0.2118}{0.0044} & 0.0273  & 1.6962 \\ 
    GPT2$_{\text{AR}}$ & \std{0.7919}{0.0319} & \std{0.1585}{0.0121} & \std{0.2515}{0.0434} & 0.0010 & 0.2857 \\ 
    Qwen  & 0.7726 & 0.1185 & 0.1860 & 0.2500 & 2.8025 \\ 
    \bottomrule
\end{tabular}
\end{adjustbox}
\end{table}

\begin{table}[H]
\caption{Performance on decompensation prediction with MC-MED dataset.}
\label{table:mcmed-decomp}
\vspace{3mm}
\begin{adjustbox}{center,max width=\textwidth}
\begin{tabular}{c|ccc|ccc}
    \toprule
    \textbf{Method} & AUROC $\uparrow$ & AUPRC $\uparrow$ & F1 $\uparrow$ & Stability $\downarrow$ & Flip $\downarrow$ \\
    \midrule
    XGB   & \std{0.6855}{0.0053} & \std{0.2462}{0.0068} & \std{0.3208}{0.0053} & 0.0452 & 0.8475 \\ 
    RF    & \std{0.6765}{0.0059} & \std{0.2364}{0.0060} & \std{0.2931}{0.0076} & 0.0258 & 0.9428 \\ 
    Mamba & \std{0.7870}{0.0293} & \std{0.4311}{0.0151} & \std{0.3652}{0.0574} & 0.0795  & 1.0265 \\ 
    GPT2  & \std{0.6978}{0.0065} & \std{0.2900}{0.0094} & \std{0.3235}{0.0030} & 0.0181 & 1.1088 \\ 
    GPT2$_{\text{AR}}$ & \std{0.6514}{0.0140} & \std{0.2418}{0.0066} & \std{0.2670}{0.0366} & 0.1025  & 2.2293 \\ 
    Qwen  & 0.5508 & 0.1496 & 0.2332 & 0.1559 & 5.0000 \\ 
    \bottomrule
\end{tabular}
\end{adjustbox}
\end{table}

\begin{table}[ht]
\caption{Performance on sepsis prediction with MC-MED dataset.}
\label{table:mcmed-sepsis}
\vspace{3mm}
\begin{adjustbox}{center,max width=\textwidth}
\begin{tabular}{c|ccc|ccc}
    \toprule
    \textbf{Method} & AUROC $\uparrow$ & AUPRC $\uparrow$ & F1 $\uparrow$ & Stability $\downarrow$ & Flip $\downarrow$ \\
    \midrule
    XGB   & \std{0.6893}{0.0} & \std{0.0893}{0.0230} & \std{0.1190}{0.0257} & 0.0494 & 0.9796 \\ 
    RF    & \std{0.6935}{0.0212} & \std{0.0866}{0.0160} & \std{0.0612}{0.0270} & 0.0202 & 0.9902 \\ 
    Mamba & \std{0.6143}{0.0077} & \std{0.0682}{0.0115} & \std{0.0863}{0.0397} & 0.0420 & 0.7500 \\ 
    GPT2  & \std{0.6437}{0.0233} & \std{0.0832}{0.0018} & \std{0.1433}{0.0101} & 0.0151 & 1.1000 \\ 
    GPT2$_{\text{AR}}$ & \std{0.6583}{0.0155} & \std{0.0835}{0.0153} & \std{0.0659}{0.0150} & 0.0001& 1.1500 \\ 
    Qwen  & 0.5508 & 0.1496 & 0.2332 & 0.1375 & 2.9440 \\ 
    \bottomrule
\end{tabular}
\end{adjustbox}
\end{table}

\begin{table}[H]
\caption{Performance on ICU transfer prediction with MIMIC-IV dataset.}
\label{table:icu}
\vspace{3mm}
\begin{adjustbox}{center,max width=\textwidth}
\begin{tabular}{c|ccc|ccc}
    \toprule
    \textbf{Method} & AUROC $\uparrow$ & AUPRC $\uparrow$ & F1 $\uparrow$ & Stability $\downarrow$ & Flip $\downarrow$ \\
    \midrule
    XGB   & \std{0.8869}{0.0060} & \std{0.5087}{0.0167} & \std{0.4371}{0.0090} & 0.0006  & 0.7572 \\ 
    RF    & \std{0.8681}{0.0054} & \std{0.4351}{0.0135} & \std{0.4265}{0.1030} & 0.0191  & 0.7867 \\
    Mamba & \std{0.8707}{0.0015} & \std{0.4834}{0.0097} & \std{0.3551}{0.0376} & 0.0621 & 1.6195 \\
    GPT2  & \std{0.8875}{0.0012} & \std{0.5353}{0.0048} & \std{0.3648}{0.0427} & 0.0632 & 1.6187 \\
    GPT2$_{\text{AR}}$ & \std{0.8598}{0.0115} & \std{0.4596}{0.0224} & \std{0.3678}{0.0148} & 0.0001  & 1.0000 \\
    Qwen  & 0.7515 & 0.5565 & 0.5587 & 0.1636 & 1.3675 \\
    \bottomrule
\end{tabular}
\end{adjustbox}
\end{table}

\begin{table}[H]
\caption{Performance on in-hospital mortality prediction with MIMIC-IV dataset.}
\label{table:ihm}
\vspace{3mm}
\begin{adjustbox}{center,max width=\textwidth}
\begin{tabular}{c|ccc|ccc}
    \toprule
    \textbf{Method} & AUROC $\uparrow$ & AUPRC $\uparrow$ & F1 $\uparrow$ & Stability $\downarrow$ & Flip $\downarrow$ \\
    \midrule
    XGB   & \std{0.8116}{0.0301} & \std{0.3282}{0.0420} & \std{0.3000}{0.0506} & 0.0104  & 0.7120 \\ 
    RF    & \std{0.8068}{0.0132} & \std{0.2817}{0.0492} & \std{0.0000}{0.0000}       & 0.0197  & 0.7823 \\ 
    Mamba & \std{0.7870}{0.0192}  & \std{0.4311}{0.0083} & \std{0.3652}{0.0010}  & 0.0276  & 1.2919 \\ 
    GPT2  & \std{0.7600}{0.0195} & \std{0.3804}{0.0225} & \std{0.2874}{0.0117} & 0.0320  & 2.7919 \\ 
    GPT2$_{\text{AR}}$ & \std{0.7107}{0.0113} & \std{0.3630}{0.0706} & \std{0.2886}{0.0101} & 0.0001 & 1.000 \\ 
    Qwen  & 0.8608 & 0.5849 & 0.5786 & 0.3035  & 1.6934 \\ 
    \bottomrule
\end{tabular}
\end{adjustbox}
\end{table}

\subsection{LLM Prompts}
\label{sec:appendix-prompts}
We provide an abridged version of the LLM prompt and a sample response in Figure \ref{fig:llm-example}.

\section{License} \label{sec:license}
Our benchmarks are derived from MC-MED-1.0.0 \cite{mc-med}, MIMIC-IV-3.1 \cite{mimic-iv}, and EHRShot-3.2 \cite{wornow2023ehrshot}. 
MC-MED and MIMIC-IV are under PhysioNet Credentialed Health Data License 1.5.0. 
We additionally adapt benchmarks from repositories: mimic-3 benchmarks (MIT) \cite{mimic3-benchmarks} and MIMIC-IV-ED benchmarks (GNU) \cite{mimic-iv-ed}.

\newpage
\definecolor{promptbg}{RGB}{245,245,250}
\definecolor{responsebg}{RGB}{240,248,255}
\definecolor{bordercol}{RGB}{220,220,235}

\newtcolorbox{llmprompt}{
  enhanced,
  colback=promptbg,
  colframe=bordercol,
  arc=2mm,
  boxrule=0.5pt,
  leftrule=3pt,
  width=\textwidth+2cm,
  enlarge left by=-1cm,
  enlarge right by=-1cm,
  title={\textbf{\textcolor{black}{Prompt}}},
  attach boxed title to top left={yshift=-2mm, xshift=4mm},
  boxed title style={colback=promptbg, colframe=bordercol, size=small, boxrule=0.5pt},
  breakable
}

\newtcolorbox{llmresponse}{
  enhanced,
  colback=responsebg,
  colframe=bordercol,
  arc=2mm,
  boxrule=0.5pt,
  leftrule=3pt,
  width=\textwidth+2cm,
  enlarge left by=-1cm,
  enlarge right by=-1cm,
  title={\textbf{\textcolor{black}{Response}}},
  attach boxed title to top left={yshift=-2mm, xshift=4mm},
  boxed title style={colback=responsebg, colframe=bordercol, size=small, boxrule=0.5pt},
  breakable
}

\begin{figure}[ht]
  \centering
  \scalebox{0.94}{%
  \begin{minipage}{\linewidth}
    \begin{llmprompt}
You are a medical AI assistant focused on the early detection of sepsis in Emergency Department (ED) patients. Your primary task is to analyze provided Electronic Health Record (EHR) data to assess the likelihood of sepsis development within the next 90 minutes from the last recorded data point. \\

The input will be a chronologically ordered sequence of real-time patient data from ED admission. Each entry follows the format: $\langle \text{time\_in\_mins} \rangle$: $\langle \text{data} \rangle$, where $\langle \text{time\_in\_mins} \rangle$ denotes minutes since admission. The data will include patient demographics, triage vitals, subsequent real-time vitals, laboratory results, and medications.

Pay close attention to trends and critical changes in the following parameters, as they relate to general sepsis indicators:
\begin{itemize}
  \item Vital Signs: Heart Rate (HR), Respiratory Rate (RR), Blood Pressure (BP, MAP), Temperature, Oxygen Saturation (SpO2).
  \item Laboratory Results: White Blood Cell Count (WBC), Lactate levels, Creatinine, Bilirubin, Platelet Count.
  \item Clinical Assessment: Altered Mental Status (e.g., changes in Glasgow Coma Scale - GCS), evidence of organ hypoperfusion (e.g., skin changes, urine output).
  \item Medications: Administration of vasopressors, intravenous fluids, and antibiotics.
\end{itemize}

To arrive at your prediction, follow these reasoning steps:
\begin{itemize}
  \item Establish Patient Context: ...
  \item eSOFA Component Assessment: ...
  \item Risk Synthesis \& Prediction: ...
\end{itemize}

Your final output MUST strictly adhere to the following format: \\
Reasoning: 2-3 paragraphs based on the above steps \\
Prediction: Yes/No \\
Risk: [0, 1)

'Yes' indicates a high risk of sepsis development within the next 90 minutes, 'No' indicates a low risk, and 'Risk' is a numerical value between 0 (no risk) and 1 (high risk). For example, "Prediction: Yes, Risk: 0.95", indicates a high risk of sepsis. Base your analysis solely on the provided EHR text and your embedded medical knowledge. \\

\textbf{Patient Info} \\
Age: 54,  Race: Asian, Sex: Female. Arrived by self. Triage vitals: temperature of 36.7$^\circ$C, heart rate of 96 bpm, respiratory rate of 16 bpm, blood pressure of 132/74 mmHg ....  \\

\textit{Timeline of Events (Minutes since Admission):} \\
119.0 min: Normosol -R IV Bolus - 500 ML administered. \\ 
151.0 min: 
\begin{itemize}
\vspace{-0.5em}
    \item Hematocrit (HCT): 40.4\%
    \item White Blood Cells (WBC): 6.1 K/uL
    \item Platelet Count (PLT): 148 K/uL
\end{itemize}
173.0 min: Vanomycin-Water Inject (PEG) 1 GRAM/200 ML IV PGBK administered. \\
177.0 min: 
\vspace{-0.5em}
\begin{itemize}
    \item Blood Urea Nitrogen (BUN): 20 mg/dL 
    \item Creatinine: 0.89 mg/dL  
\end{itemize}
....
    \end{llmprompt}

    \vspace{1mm}

    \begin{llmresponse}
The patient is a 54-year-old Asian female who presented to the ED with a chief complaint of "abnormal lab" and triaged as urgent (acuity 3). Her initial vitals include a normal temperature (36.7$^\circ$C), tachycardia (HR 96), and otherwise unremarkable respiratory and blood pressure parameters. No overt signs of infection or sepsis are immediately evident from the triage vitals .... \textbf{Prediction: No, Risk: 0.25}
    \end{llmresponse}
  \end{minipage}
  }
  \caption{Example of an LLM prompt-response pair for early event prediction of sepsis onset on the MC-MED dataset.}
  \label{fig:llm-example}
\end{figure}


\end{document}